\PassOptionsToPackage{numbers}{natbib}
\documentclass{article}

\usepackage[dblblindworkshop, final]{neurips_2025} 
\workshoptitle{MATH-AI}
\usepackage{amsmath}
\usepackage{graphicx}
\usepackage{placeins} 
\usepackage{booktabs}

\usepackage[utf8]{inputenc} 
\usepackage[T1]{fontenc}    
\usepackage{hyperref}       
\usepackage{url}            
\usepackage{booktabs}       
\usepackage{amsfonts}       
\usepackage{nicefrac}       
\usepackage{microtype}      
\usepackage{xcolor}         

\begin{document}
\title{Concept Generalization in Humans and Large Language Models: Insights from the Number Game}

\author{%
  Arghavan Bazigaran \\
  Department of Biomedical Engineering, \\ 
  Sungkyunkwan University, South Korea \\
  \texttt{a.bazigaran@skku.edu} \\
  \And
  Hansem Sohn* \\
  Center for Neuroscience Imaging Research, \\ 
  Institute for Basic Science, \\ 
  Department of Biomedical Engineering, \\ 
  Sungkyunkwan University, South Korea \\
  * Correspondence: \texttt{hansem@skku.edu} \\
}
\maketitle

\begin{abstract}

  We compare human and large language model (LLM) generalization in the number game, a concept inference task. Using a Bayesian model as an analytical framework, we examined the inductive biases and inference strategies of humans and LLMs. The Bayesian model captured human behavior better than LLMs in that humans flexibly infer rule-based and similarity-based concepts, whereas LLMs rely more on mathematical rules. Humans also demonstrated a few-shot generalization, even from a single example, while LLMs required more samples to generalize. These contrasts highlight the fundamental differences in how humans and LLMs infer and generalize mathematical concepts.
\end{abstract}

\section{Introduction}

Large language models (LLMs) have been evaluated on various mathematical reasoning benchmarks\cite{Cobbe2021-gz,hendrycks2021measuring,Glazer2024-lj,lewkowycz2022solving,sessler2024benchmarking}. Among these, logical deduction from limited information has remained a particularly challenging domain for LLMs \cite{lu2023mathvista,Saparov2022-wm,Yue:EECS-2025-121}. To address this challenge, recent studies have moved beyond final-answer evaluation to analyze intermediate reasoning steps \cite{Saparov2022-wm,Xia2024-em,Mirzadeh2024-wn,gao2023pal} and combined measures of overall accuracy, stepwise soundness, and representational-level analysis \cite{Zhou2025-ra}.

Here, we use the number game\cite{Tenenbaum1999-xf} to examine how LLMs reason about mathematical concepts and generalize from a few examples. Unlike prior studies embedding reasoning in complex visual or linguistic contexts \cite{lu2023mathvista,Xia2024-em,Zhou2025-ra,Cobbe2021-gz}, the number game allows us to systematically probe inductive biases and inference strategies in a few-shot setting. This task has revealed that humans possess rich number concepts based on either mathematical rules (e.g., even numbers) or similarity (e.g., 11 is closer to 10 than 15) and can flexibly infer and generalize these concepts\cite{Bigelow2016-mp}. These computational processes were modeled by a Bayesian framework\cite{sohn2019bayesian,sohn2021neural,sohn2021validating} that incorporates the number concepts in its prior and integrates them with observed data to support flexible generalization.

In this work, We first test whether LLMs infer number concepts along the rule–similarity continuum based on limited information in the number game. We then compare LLMs' behavior with human data \cite{Bigelow2016-mp} and with a Bayesian model that provides a normative prediction of generalization behavior. We further fit the Bayesian model to capture each agent's inductive bias toward rules or similarity and perform ablation analyses on model components to identify the source of human-LLM differences.
Together, our work provides insights into LLMs’ mathematical reasoning and how it diverges from human cognition.

\section{Methods}

\subsection{Human and LLM Dataset}

 We used a publicly available dataset of human behavior \cite{Bigelow2016-mp} from a structured concept inference task known as the number game \cite{Tenenbaum1999-xf}. In each trial, participants were shown a set of example numbers (e.g., \{2, 4, 6\}) and judged whether a given target number (1–100) belonged to the same underlying mathematical concept. Binary (yes/no) responses from at least nine subjects were averaged to obtain $p(\text{yes})$ per target number
 (see appendix \ref{appendix:dataset_structure} for dataset details).

\begin{figure}[htbp]
    \centering
    \includegraphics[width=0.9\linewidth]{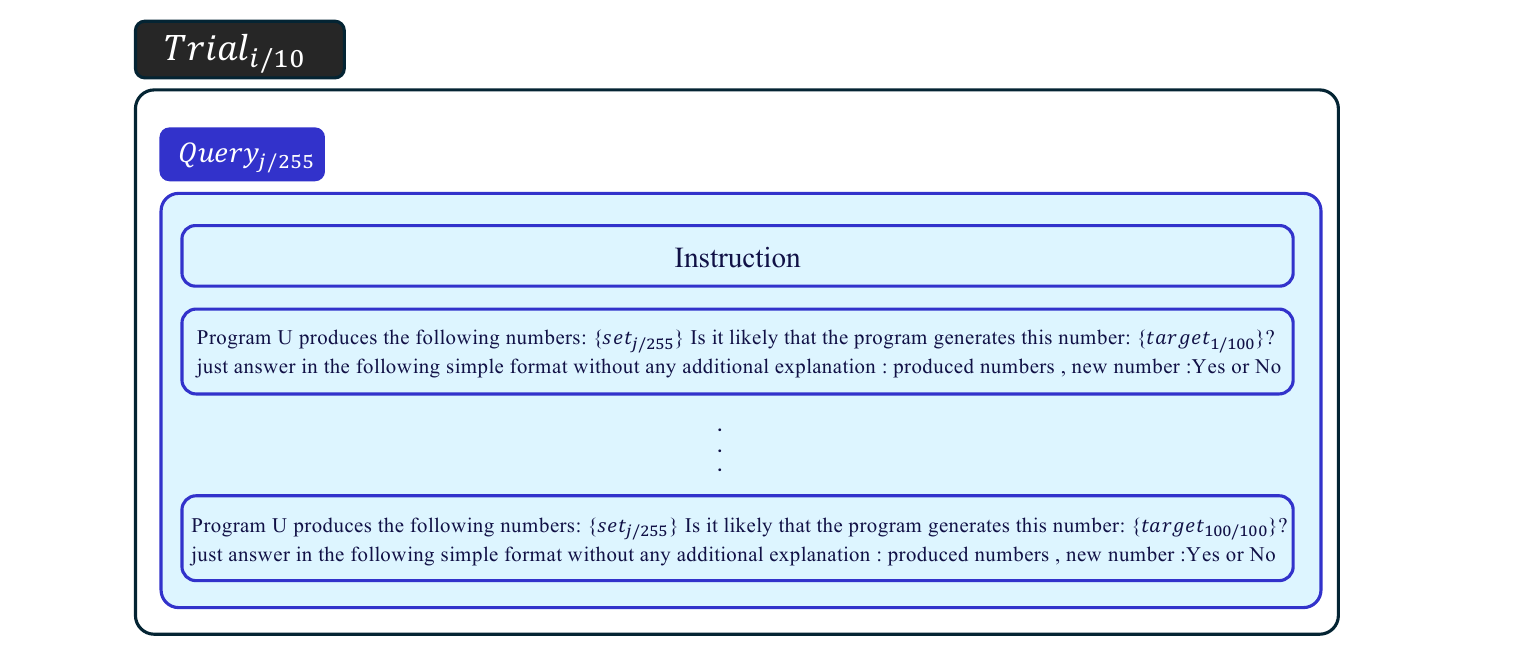}
    \caption{Task structure used for LLM evaluation. For each example set, the model is queried on all possible targets, and this process is repeated across trials.
    }
    \label{fig:prompt}
\end{figure}

We attempted to replicate the human experiment \cite{Bigelow2016-mp} as closely as possible. To do so, we prompted GPT-o1-mini(OpenAI) via a public API with 255 example sets, querying all target numbers from 1 to 100 (Figure \ref{fig:prompt}; see section \ref{other LLMs} for other LLMs). To ensure consistency, we included minimal instructions encouraging direct yes/no responses and averaged the results over 10 trials.

\subsection{Bayesian Model}

To model the inference process in the number game, we adopted the Bayesian model \cite{Tenenbaum1999-xf,Bigelow2016-ma}, which evaluates a hypothesis space of possible numerical concepts given a set of example numbers. The model has three components. First, each hypothesis is assigned a prior probability $P(h)$, reflecting its plausibility before observing any data. The hypothesis space includes both rule-based (e.g., even numbers, multiples, and primes) and similarity-based concepts formed by all possible intervals within 1–100 (e.g., $20$–$30$, $25$–$75$, ${7}$). Second, given examples $X$, the model computes a likelihood $P(X \mid h)$, reflecting how well a hypothesis explains the examples. The likelihood incorporates the \textit{size principle}, which favors more specific concepts with a smaller cardinality (e.g., power of 2 over even numbers for \{2, 8\}). Prior and likelihood are combined via Bayes’ rule to produce a posterior:

$P(h \mid X) = \frac{P(X \mid h) \cdot P(h)}{\sum_{h'} P(X \mid h') \cdot P(h')}$

The model generalizes to a new target $y$ by averaging predictions $P(y\mid h)$ across hypotheses weighted by their posterior probabilities $P(h\mid X)$ (\textit{hypothesis averaging}). Thus, the full model integrates priors for rule-based and interval-based concepts, likelihood with the size principle, and hypothesis averaging for generalization (appendix \ref{Appendix:Bayesian Model Details}).

\subsection{Model Fitting and Evaluation}

Bayesian models served a dual role in our study: first, as a benchmark for comparing humans and LLMs, and second, as a diagnostic tool for understanding LLM's behavior. We first assessed LLM behavior using a baseline Bayesian model \cite{Tenenbaum1999-xf,Bigelow2016-ma} previously shown to capture human behavior, thereby providing a benchmark for comparison. A key parameter was the rule bias ($\lambda$), reflecting the model’s prior preference for rule-based over interval-based concepts. We also treated $\lambda$ as a free parameter and compared fitted values between humans and LLM to gauge their rule bias. 

To probe structural contributions, we built three model variants (Figure \ref{fig:BP_joint}C). \textit{Binary Likelihood ($BinL$) model} replaces the size-based likelihood with a binary likelihood that considers only whether the examples conform to the concept. \textit{Maximum A Posteriori ($MAP$) model} replaces hypothesis averaging with selection of the single hypothesis that maximizes the posterior. \textit{Maximum Likelihood ($MaxL$) model} removes priors and instead predicts by maximizing the likelihood. These variants allow us to test the roles of size-based likelihood, hypothesis averaging, and prior in explaining human and LLM behavior.

Model evaluation used Jensen-Shannon Divergence (JSD), a bounded measure (0–1) quantifying (dis)similarity between predicted and observed response distributions. The smaller the JSD, the more similar the two distributions are. JSD was computed per example set and averaged across sets to assess the overall similarity. 

\section{Results}

In general, GPT O1-mini (hereafter GPT) showed diverse and rich response patterns for the number game with a signature of both similarity- and rule-based generalization (section \ref{other LLMs} for other LLM).

\subsection{Qualitative comparisons: no one-shot generalization in GPT}

\begin{figure}[htbp]
    \centering
    \includegraphics[width=1\linewidth]{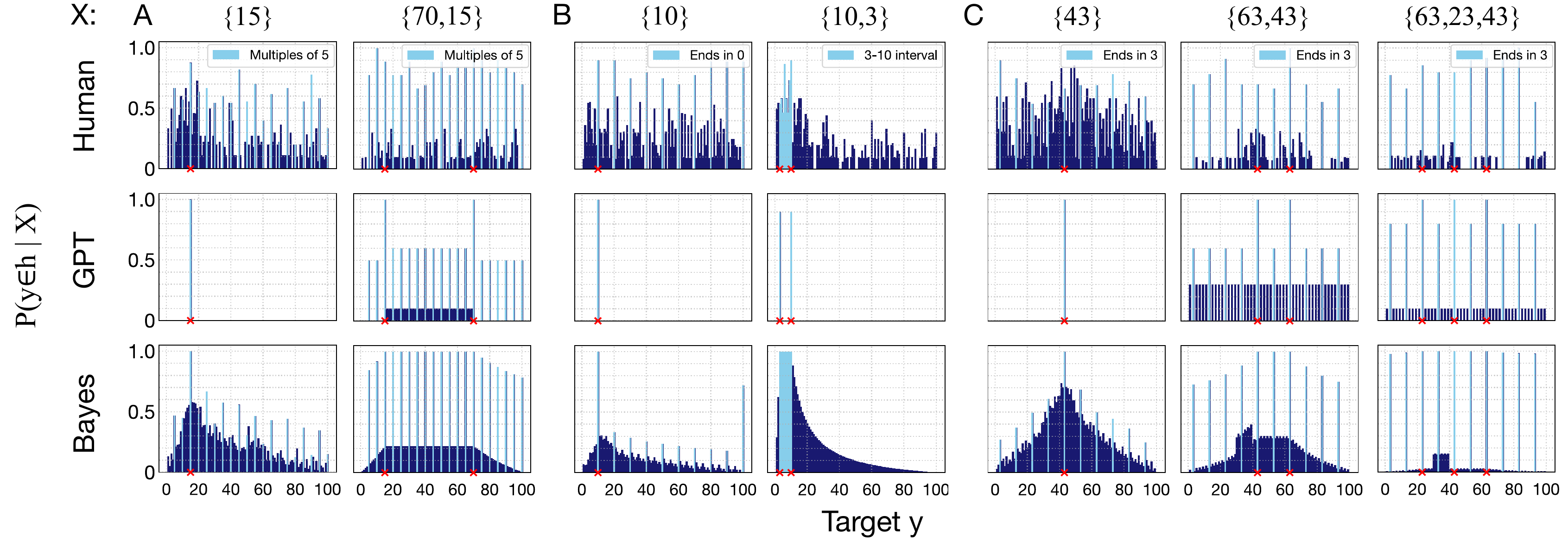}
    \caption{
    Number-game behavior of human, GPT, and baseline Bayesian model (Bayes) for selected example sets (top titles), showing the probability that a target number $y$ belongs to the same concept $h$ as the given example set $X$. Part of the data consistent with a mathematical rule is shown in pale blue.
    }
    \label{fig:example_bar}
\end{figure}

Figure\ref{fig:example_bar} compares humans, GPT, and the baseline Bayesian model ($\lambda$=0.5 rule bias) across representative example sets. The most salient feature of GPT's behavior is its response to a single example ({15},{10},{43}): GPT remains anchored to the given number while both humans and the Bayesian model show a mixture of interval-based (responses concentrated near the example) and rule-based (sparse peaks like 'multiples of 5') generalization.

Adding a second example shifts the responses differently across agents. When the examples are close (B: ${10,3}$), humans and the Bayesian model rely more on interval-based concepts, as indicated by the gradually decaying ratings from the examples. When two examples are far apart (A: ${70,15}$), all agents identify a rule consistent with both examples with weaker interval-based patterns.

With three examples (c), rule-based concepts become eminent across all agents. These suggest that while GPT relies on both rule and similarity, it does not generalize from a single sample, unlike humans and the Bayesian model. Overall, the Bayesian model aligns more closely with human responses than GPT, capturing one-shot generalization and flexible use of rule and interval concepts.

\subsection{Quantitative comparisons: baseline Bayes captures human behavior better than GPT }

We next quantified human-GPT-Bayes differences across all example sets using mean pairwise JSD as a metric (Table\ref{tab:jsd-comparison}, Figure\ref{fig:BP_joint}A). Overall, divergence was lower for human-Bayes than GPT-human and GPT-Bayes, indicating that human responses align better with the Bayesian model than with GPT. Consistent with the qualitative observations above, this result suggests that GPT employs a different generalization strategy from humans and the Bayesian model, particularly in the one-example trials.

\subsection{Set-length effect: Poor sample efficiency in GPT}

\begin{figure}[htbp]
    \centering
    \includegraphics[width=1\linewidth]{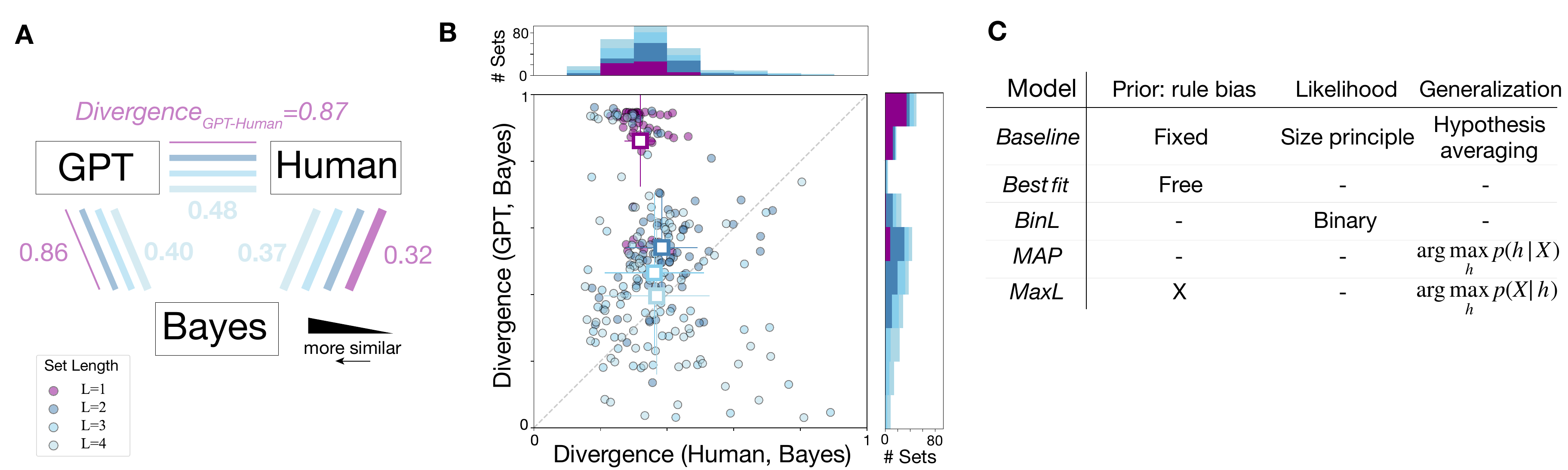}
    \caption{
    (A) Quantitative comparison of humans, GPT, and baseline Bayes. Mean divergence for each set length is shown for human–Bayes–GPT comparisons. (B) The central plot shows the per-set divergence: GPT-Bayes (y-axis) and Human-Bayes (x-axis). Marginal histograms show the value distributions. Error bars show the standard deviation. (C) Summary of the Bayesian models. The baseline, best fit and variants; BinL (binary likelihood), MAP (maximum a posteriori), and MaxL (maximum likelihood).
    }
    \label{fig:BP_joint}
\end{figure}

To dissect the human-GPT difference in one-example generalization, we examined per-set divergences from the baseline Bayesian model to explore how the behavioral gap changes with the number of examples. Figure \ref{fig:BP_joint}B shows per-set divergence distributions: GPT-Bayes alignment is generally worse, particularly for single-example sets where GPT rarely generalizes beyond the given example (Figure \ref{fig:example_bar}). GPT-Bayes alignment improves as the number of examples increases, and with $L=4$, the GPT-Bayes divergence becomes comparable to the human-Bayes divergence. In contrast, the human-Bayes divergence is largely unaffected by set length. This pattern indicates poorer sample efficiency in GPT: humans exhibit one-shot generalization using rich concepts based on rule and similarity, whereas GPT requires multiple examples. 

\subsection{Bayesian model fitting: higher rule bias in GPT}

Building on the baseline model analysis that benchmarked GPT's behavior against human and Bayesian models, we sought to identify the sources of GPT's poor generalization using Bayesian modeling as a diagonostic tool. We hypothesized that one possible reason for the GPT-human difference is that GPT lacks the rich and complex number concepts that humans possess. To test this, we next fit the Bayesian model to compare the relative reliance on rule-based and interval-based concepts. Specifically, we examined the rule bias ($\lambda$), the relative probability of using rule-based over interval-based concepts, by fitting the $\lambda$ parameter through a grid search. This fitting improved alignment for both humans and GPT compared to the baseline model ($\lambda=0.5$). The best-fit value of $\lambda$ was 0.9 in humans and 1 in GPT. This higher rule bias in GPT is consistent with its limited reliance on intervals and its lack of similarity gradient seen in humans and Bayes (Figure\ref{fig:example_bar}A, B).

\subsection{Nested model comparison: maximum likelihood inference in GPT}

Having identified rule-similarity differences through the fitted rule-bias parameter, we next examined which other structural components of the Bayesian model are critical for explaining GPT's behaviors beyond its number concepts. In the Bayesian framework, number-concept composition is only one component that may account for GPT's distinct generalization patterns. The human-GPT difference could also arise from other elements such as likelihood computation via the size principle, inference steps based on posterior estimation, or generalization strategy with hypothesis averaging. We tested whether GPT’s weaker sample efficiency reflects differences in any of these computational structures. 
To this end, we built three model variants of the full model, each isolating a core computational component: $BinL$, $MAP$ and $MaxL$ (Table\ref{tab:jsd-comparison}). For humans, the best and worst variants are $BinL$ and $MaxL$, respectively, highlighting the contribution of the Bayesian prior and hypothesis averaging. The opposite pattern is observed for GPT (best: $MaxL$, worst: $BinL$), suggesting that GPT's behavior depends more strongly on the likelihood-based inference than on the prior structure. When we examine the fit by example sets (Figure\ref{fig:category_merged}), the $MaxL$ model shows a substantial improved fit for the $L=1$ sets. This result implies that GPT's lack of one-shot generalization arises from its heavy dependence on the maximum likelihood inference strategy.

\begin{table}[ht]
  \caption{Mean divergence of model variants (mean ± SEM). 
  Left: full Bayesian models. Right: fixed-parameter variants.}
  \label{tab:jsd-comparison}
  \centering
  \small
  \begin{minipage}{0.48\textwidth}
    \centering
    \begin{tabular}{lcc}
      \toprule
      & Baseline & Best Fit \\
      \midrule
      Human–Bayes & $0.36 \pm 0.01$ & $\mathbf{0.30 \pm 0.01}$ \\
      GPT–Bayes   & $0.56 \pm 0.02$ & $\mathbf{0.49 \pm 0.02}$ \\
      \bottomrule
    \end{tabular}
  \end{minipage}
  \hspace{0.02\textwidth}
  \begin{minipage}{0.45\textwidth}
    \centering
    \begin{tabular}{@{}ccc@{}}
      \toprule
      $MAP$ & $BinL$ & $MaxL$ \\
      \midrule
      $0.60 \pm 0.01$ & $\mathbf{0.33 \pm 0.01}$ & $0.66 \pm 0.01$ \\
      $0.50 \pm 0.02$ & $0.65 \pm 0.01$ & $\mathbf{0.42 \pm 0.02}$ \\
      \bottomrule
    \end{tabular}
  \end{minipage}
\end{table}

\subsection{Other LLMs in  the number game: less structured behavior} \label{other LLMs}
We also tested several other LLMs, including GPT-4o (OpenAI), DeepSeek-V3, and LLaMA-3-8B (Meta), to examine whether other models exhibit more human-like generalization patterns or reveal novel reasoning strategies absent in GPT-o1-mini. Unlike GPT O1-mini, these models showed weaker generalization and failed to produce rich, systematic patterns of rule-based and interval-based concepts (\ref{fig:example-llm} for representative examples). These results suggest that differences in model architecture and training can shape the formation of number concepts, as well as the reasoning and generalization behaviors that emerge from them.

\section{Discussion}

Recent progress in LLMs' mathematical reasoning benchmarks poses important questions: Can they generalize mathematical concepts across contexts? How closely do their internal representations resemble those of humans?\cite{binz2023turning,asperti2025thinking} Using the number game and Bayesian modeling as an interpretable scaffold, we probed rule bias, sample efficiency, and generalization strategies in humans and LLMs.

Overall, human behavior was better predicted by the Bayesian model than GPT, in that human generalization strategy depends on the rich structure of the Bayesian prior and the averaging of multiple hypotheses for prediction. This human-Bayes alignment is not surprising given the model's flexibility to capture near-optimal behaviors across different domains\cite{sohn2019bayesian,sohn2021neural,sohn2021validating,sohn2013dichotomy}. While humans showed signature of rich number concepts, GPT relies heavily on rule-based inference, showing little evidence of similarity-based generalization. This suggests that GPT has an inductive bias toward rule-based inference in mathematical reasoning and that its numerical sense lacks the full development of similarity-based concepts observed in humans. Because hypothesis space in the Bayesian model's prior includes only primitive mathematical rules, it remains to be tested whether other rule-based concepts may better account for GPT's behavior such as compositionally structured rules used in language \cite{Bigelow2016-ma}(in appendix \ref{appendix:dataset_structure}). Future work should also assess more recent LLMs to track how model architecture and training affect number concepts and generalization behavior. 

Another key finding was that GPT mostly fails to generalize beyond the example when only one sample is given. This was in stark contrast to the humans' one-shot generalization, which elicited both rule-based and similarity-based inferences. The systematic ablation of model components further reveals that GPT's poorer sample efficiency arises from its maximum-likelihood inference strategy. One possibility is that GPT may not consider multiple hypotheses during an inference step due to limited memory capacity—a possibility that warrants further investigation in future studies.

Together, these findings highlight fundamental differences between human and LLM mathematical reasoning. While GPT can infer a variety of concepts, its in-context generalization remains limited, requiring multiple examples and lacking the flexible numerical sense observed in humans. This limitation may stem from the fact that LLMs are trained predominantly on next-token prediction\cite{gloeckle2024better} using natural language. Future work could explore different training data that incorporate mathematical and symbolic representations\cite{lu2024mathcoder2,lewkowycz2022solving}, prompt-engineering methods for better in-context inference, or targeted fine-tuning curricula shown to enhance numerical understanding\cite{yang2024number} to enrich LLMs' inductive biases. Such approaches may help bridge the gap in sample efficiency and enable richer, human-like generalization of numerical concepts.

\vspace*{\fill}
\newpage

\bibliography{references}

@article{sohn2019bayesian,
  title={Bayesian computation through cortical latent dynamics},
  author={Sohn, Hansem and Narain, Devika and Meirhaeghe, Nicolas and Jazayeri, Mehrdad},
  journal={Neuron},
  volume={103},
  number={5},
  pages={934--947},
  year={2019},
  publisher={Elsevier}
}

@article{sohn2021neural,
  title={Neural implementations of Bayesian inference},
  author={Sohn, Hansem and Narain, Devika},
  journal={Current Opinion in Neurobiology},
  volume={70},
  pages={121--129},
  year={2021},
  publisher={Elsevier}
}

@article{sohn2013dichotomy,
  title={Dichotomy in perceptual learning of interval timing: calibration of mean accuracy and precision differ in specificity and time course},
  author={Sohn, Hansem and Lee, Sang-Hun},
  journal={Journal of neurophysiology},
  volume={109},
  number={2},
  pages={344--362},
  year={2013},
  publisher={American Physiological Society Bethesda, MD}
}

@article{sohn2021validating,
  title={Validating model-based Bayesian integration using prior--cost metamers},
  author={Sohn, Hansem and Jazayeri, Mehrdad},
  journal={Proceedings of the National Academy of Sciences},
  volume={118},
  number={25},
  pages={e2021531118},
  year={2021},
  publisher={National Academy of Sciences}
}

@ARTICLE{lu2023mathvista,
  title={Mathvista: Evaluating mathematical reasoning of foundation models in visual contexts},
  author={Lu, Pan and Bansal, Hritik and Xia, Tony and Liu, Jiacheng and Li, Chunyuan and Hajishirzi, Hannaneh and Cheng, Hao and Chang, Kai-Wei and Galley, Michel and Gao, Jianfeng},
  journal={arXiv preprint arXiv:2310.02255},
  year={2023}
}

@ARTICLE{Zhou2025-ra,
  title         = "Dissecting logical reasoning in {LLMs}: A fine-grained
                   evaluation and supervision study",
  author        = "Zhou, Yujun and Ye, Jiayi and Ling, Zipeng and Han, Yufei and
                   Huang, Yue and Zhuang, Haomin and Liang, Zhenwen and Guo,
                   Kehan and Guo, Taicheng and Wang, Xiangqi and Zhang,
                   Xiangliang",
  journal       = "arXiv [cs.CL]",
  month         =  jun,
  year          =  2025,
  archivePrefix = "arXiv",
  primaryClass  = "cs.CL"
}

@ARTICLE{Saparov2022-wm,
  title         = "Language models are greedy reasoners: A systematic formal
                   analysis of chain-of-thought",
  author        = "Saparov, Abulhair and He, He",
  journal       = "arXiv [cs.CL]",
  month         =  oct,
  year          =  2022,
  archivePrefix = "arXiv",
  primaryClass  = "cs.CL"
}

@ARTICLE{Cobbe2021-gz,
  title         = "Training verifiers to solve math word problems",
  author        = "Cobbe, Karl and Kosaraju, Vineet and Bavarian, Mohammad and
                   Chen, Mark and Jun, Heewoo and Kaiser, Lukasz and Plappert,
                   Matthias and Tworek, Jerry and Hilton, Jacob and Nakano,
                   Reiichiro and Hesse, Christopher and Schulman, John",
  journal       = "arXiv [cs.LG]",
  month         =  oct,
  year          =  2021,
  archivePrefix = "arXiv",
  primaryClass  = "cs.LG"
}

@ARTICLE{Glazer2024-lj,
  title         = "{FrontierMath}: A benchmark for evaluating advanced
                   mathematical reasoning in {AI}",
  author        = "Glazer, Elliot and Erdil, Ege and Besiroglu, Tamay and
                   Chicharro, Diego and Chen, Evan and Gunning, Alex and Olsson,
                   Caroline Falkman and Denain, Jean-Stanislas and Ho, Anson and
                   Santos, Emily de Oliveira and Järviniemi, Olli and Barnett,
                   Matthew and Sandler, Robert and Vrzala, Matej and Sevilla,
                   Jaime and Ren, Qiuyu and Pratt, Elizabeth and Levine, Lionel
                   and Barkley, Grant and Stewart, Natalie and Grechuk, Bogdan
                   and Grechuk, Tetiana and Enugandla, Shreepranav Varma and
                   Wildon, Mark",
  journal       = "arXiv [cs.AI]",
  month         =  nov,
  year          =  2024,
  archivePrefix = "arXiv",
  primaryClass  = "cs.AI"
}

@ARTICLE{Mirzadeh2024-wn,
  title         = "{GSM}-symbolic: Understanding the limitations of mathematical
                   reasoning in Large Language Models",
  author        = "Mirzadeh, Iman and Alizadeh, Keivan and Shahrokhi, Hooman and
                   Tuzel, Oncel and Bengio, Samy and Farajtabar, Mehrdad",
  journal       = "arXiv [cs.LG]",
  month         =  oct,
  year          =  2024,
  archivePrefix = "arXiv",
  primaryClass  = "cs.LG"
}

@ARTICLE{Xia2024-em,
  title         = "Evaluating mathematical reasoning beyond accuracy",
  author        = "Xia, Shijie and Li, Xuefeng and Liu, Yixin and Wu, Tongshuang
                   and Liu, Pengfei",
  journal       = "arXiv [cs.CL]",
  month         =  apr,
  year          =  2024,
  archivePrefix = "arXiv",
  primaryClass  = "cs.CL"
}

@ARTICLE{Bigelow2016-ma,
  title   = "Inferring priors in compositional cognitive models",
  author  = "Bigelow, E and Piantadosi, S",
  journal = "CogSci",
  year    =  2016
}

@ARTICLE{Bigelow2016-mp,
  title     = "A large dataset of generalization patterns in the number game",
  author    = "Bigelow, Eric and Piantadosi, Steven T",
  journal   = "J. Open Psychol. Data",
  publisher = "Ubiquity Press, Ltd.",
  volume    =  4,
  number    =  1,
  pages     =  4,
  month     =  mar,
  year      =  2016
}

@ARTICLE{Tenenbaum1999-xf,
  title   = "Rules and similarity in concept learning",
  author  = "Tenenbaum, J",
  journal = "Neural Inf Process Syst",
  pages   = "59--65",
  month   =  nov,
  year    =  1999
}

@article{gloeckle2024better,
  title={Better \& faster large language models via multi-token prediction},
  author={Gloeckle, Fabian and Idrissi, Badr Youbi and Rozi{\`e}re, Baptiste and Lopez-Paz, David and Synnaeve, Gabriel},
  journal={arXiv preprint arXiv:2404.19737},
  year={2024}
}

@mastersthesis{Yue:EECS-2025-121,
    Author= {Yue, Jonathan and Klein, Daniel},
    Title= {Benchmarking LLMs on Advanced Mathematical Reasoning},
    School= {EECS Department, University of California, Berkeley},
    Year= {2025},
    Month= {May},
    Url= {http://www2.eecs.berkeley.edu/Pubs/TechRpts/2025/EECS-2025-121.html},
    Number= {UCB/EECS-2025-121},
}

@article{sessler2024benchmarking,
  title={Benchmarking large language models for math reasoning tasks},
  author={Se{\ss}ler, Kathrin and Rong, Yao and G{\"o}zl{\"u}kl{\"u}, Emek and Kasneci, Enkelejda},
  journal={arXiv preprint arXiv:2408.10839},
  year={2024}
}

@inproceedings{gao2023pal,
  title={Pal: Program-aided language models},
  author={Gao, Luyu and Madaan, Aman and Zhou, Shuyan and Alon, Uri and Liu, Pengfei and Yang, Yiming and Callan, Jamie and Neubig, Graham},
  booktitle={International Conference on Machine Learning},
  pages={10764--10799},
  year={2023},
  organization={PMLR}
}

@article{binz2023turning,
  title={Turning large language models into cognitive models},
  author={Binz, Marcel and Schulz, Eric},
  journal={arXiv preprint arXiv:2306.03917},
  year={2023}
}

@article{hendrycks2021measuring,
  title={Measuring mathematical problem solving with the math dataset},
  author={Hendrycks, Dan and Burns, Collin and Kadavath, Saurav and Arora, Akul and Basart, Steven and Tang, Eric and Song, Dawn and Steinhardt, Jacob},
  journal={arXiv preprint arXiv:2103.03874},
  year={2021}
}

@article{lewkowycz2022solving,
  title={Solving quantitative reasoning problems with language models},
  author={Lewkowycz, Aitor and Andreassen, Anders and Dohan, David and Dyer, Ethan and Michalewski, Henryk and Ramasesh, Vinay and Slone, Ambrose and Anil, Cem and Schlag, Imanol and Gutman-Solo, Theo and others},
  journal={Advances in neural information processing systems},
  volume={35},
  pages={3843--3857},
  year={2022}
}

@article{lu2024mathcoder2,
  title={Mathcoder2: Better math reasoning from continued pretraining on model-translated mathematical code},
  author={Lu, Zimu and Zhou, Aojun and Wang, Ke and Ren, Houxing and Shi, Weikang and Pan, Junting and Zhan, Mingjie and Li, Hongsheng},
  journal={arXiv preprint arXiv:2410.08196},
  year={2024}
}

@article{asperti2025thinking,
  title={Thinking Machines: Mathematical Reasoning in the Age of LLMs},
  author={Asperti, Andrea and Naibo, Alberto and Coen, Claudio Sacerdoti},
  journal={arXiv preprint arXiv:2508.00459},
  year={2025}
}

@article{yang2024number,
  title={Number cookbook: Number understanding of language models and how to improve it},
  author={Yang, Haotong and Hu, Yi and Kang, Shijia and Lin, Zhouchen and Zhang, Muhan},
  journal={arXiv preprint arXiv:2411.03766},
  year={2024}
}
\begin{ack}
This research was supported by Basic Science Research Program through the National Research Foundation of Korea(NRF) funded by the Ministry of Education(RS-2025-25431602).
\end{ack}

\newpage
\appendix
\appendix
\renewcommand{\thesubsection}{\Alph{subsection}}
\renewcommand{\thefigure}{\thesubsection.\arabic{figure}}
\renewcommand{\thetable}{\thesubsection.\arabic{table}}
\subsection {Dataset Structure}
\setcounter{figure}{0}
\setcounter{table}{0}
\label{appendix:dataset_structure}
The dataset contains 255 unique example sets, sampled from 79 full concepts. These full concepts were generated by applying mathematical functions to a set of base number categories (e.g., all numbers, evens, odds, squares, cubes, and primes). A wide variety of concepts were constructed, with the applied functions ranging from simple identities (e.g., $f(n) = n$) to compositions of mathematical operations (e.g., $f(n) = 2^n + 1$). The final example sets include both single-item sets and subsets (up to length 4) drawn from these full concepts.

\subsection{Bayesian Model Details}
\setcounter{figure}{0}
\setcounter{table}{0}
\label{Appendix:Bayesian Model Details}
\paragraph{Bayesian Model Components}
Two core components of the Bayesian inference model were designed to capture human generalization behavior: the likelihood function and the generalization mechanism. The likelihood assumes that examples are sampled uniformly at random from a concept, leading to a preference for more specific (i.e., smaller) hypotheses that still explain the data. Furthermore, the distinction between competing hypotheses becomes sharper as the number of examples increases. This \textbf{size principle} penalize broader hypotheses and amplifies likelihood differences with more examples. The likelihood is defined as follows:

\[
P(X \mid h) =
\begin{cases}
\frac{1}{|h|^n}, & \text{if } \forall_i\, x_i \in h \\
0, & \text{otherwise}
\end{cases}
\]

where $|h|$ is the size of the hypothesis and $n$ is the number of observed examples in the set.

To generalize the underlying concept of an example set $X$ to a new target number $y$, the model considers all candidate hypotheses for $X$, assigns each a binary prediction for $y$ (1 if $y \in h, 0$ otherwise), and averages these predictions weighted by the posterior probability of each hypothesis. This approach, known as \textbf{hypothesis averaging}, is defined as:

\begin{equation*}
P(y \mid X) = \sum_{h : y \in h} P(h \mid X)
\end{equation*}

\vspace{0.5em}
\paragraph{Noise Parameter ($\alpha$).}
To account for the variability in human responses, we adopt a noisy version of the model introduced by \cite{Bigelow2016-ma}. The noise parameter $\alpha$ represents the probability of structured (concept-driven) generation, while $1 - \alpha$ is the lapse rate, reflecting the probability of unstructured, uniform responses drawn from the entire number range (1–100). With this addition, the likelihood function no longer assumes that all examples must perfectly align with the hypothesis to be considered. Instead, even when all examples fall outside the hypothesis set, the hypothesis is not completely ruled out, it is simply less probable. If $x_i \in h$, the model assumes that it could have been generated from either the concept or noise. In the other case, it assumes that the example was generated from noise alone. Formally:

\[
p(x_i \mid h)=
\begin{cases}
\alpha\,\dfrac{1}{|h|} + (1-\alpha)\,\dfrac{1}{100}, & x_i \in h,\\[6pt]
(1-\alpha)\,\dfrac{1}{100}, & x_i \notin h,
\end{cases}
\qquad
P(X\mid h)=\prod_{i=1}^{n} p(x_i\mid h).
\]

In the generalization step, if $y \notin h$, the model allows for a soft prediction via the same noise mechanism:

\[
p(y\mid h)=
\begin{cases}
1, & y\in h,\\
1-\alpha, & y\notin h,
\end{cases}
\qquad
P(y\mid X)=\sum_{h} P(h\mid X)\, p(y\mid h).
\]
\vspace{0.5em}
\paragraph{Prior Over Concepts.}
Our hypothesis space, following \cite{Bigelow2016-ma}, includes both rule-based and interval-based numerical concepts. The prior is defined as a mixture of two uniform distributions: one over rule-based concepts and one over interval-based concepts. The mixing parameter $\lambda \in [0, 1]$ determines the weight of each type, where $\lambda = 1$ indicates full rule bias and $\lambda = 0$ indicates full interval bias.

\vspace{0.5em}
\paragraph{Parameter Optimization.}
We performed a grid search over both $\alpha$ and $\lambda$ to identify the best-fit values. The grid included:
\[
\{0, 0.01, 0.3, 0.5, 0.7, 0.9, 0.99, 1\}
\]
Additional fine-tuning was conducted near the optimal points. For each parameter setting, the model was run over all example sets, and mean Jensen-Shannon divergence (JSD) was computed to evaluate the fit to human or LLM data. Optimization results revealed a stronger rule bias in GPT ($\lambda = 1$) compared to humans ($\lambda = 0.9$), as well as a higher lapse rate in human responses ($1 - \alpha = 0.15$) than in GPT ($1 - \alpha = 0$).
\begin{figure}[htbp]
                \centering
                \includegraphics[width=0.9\linewidth]
                {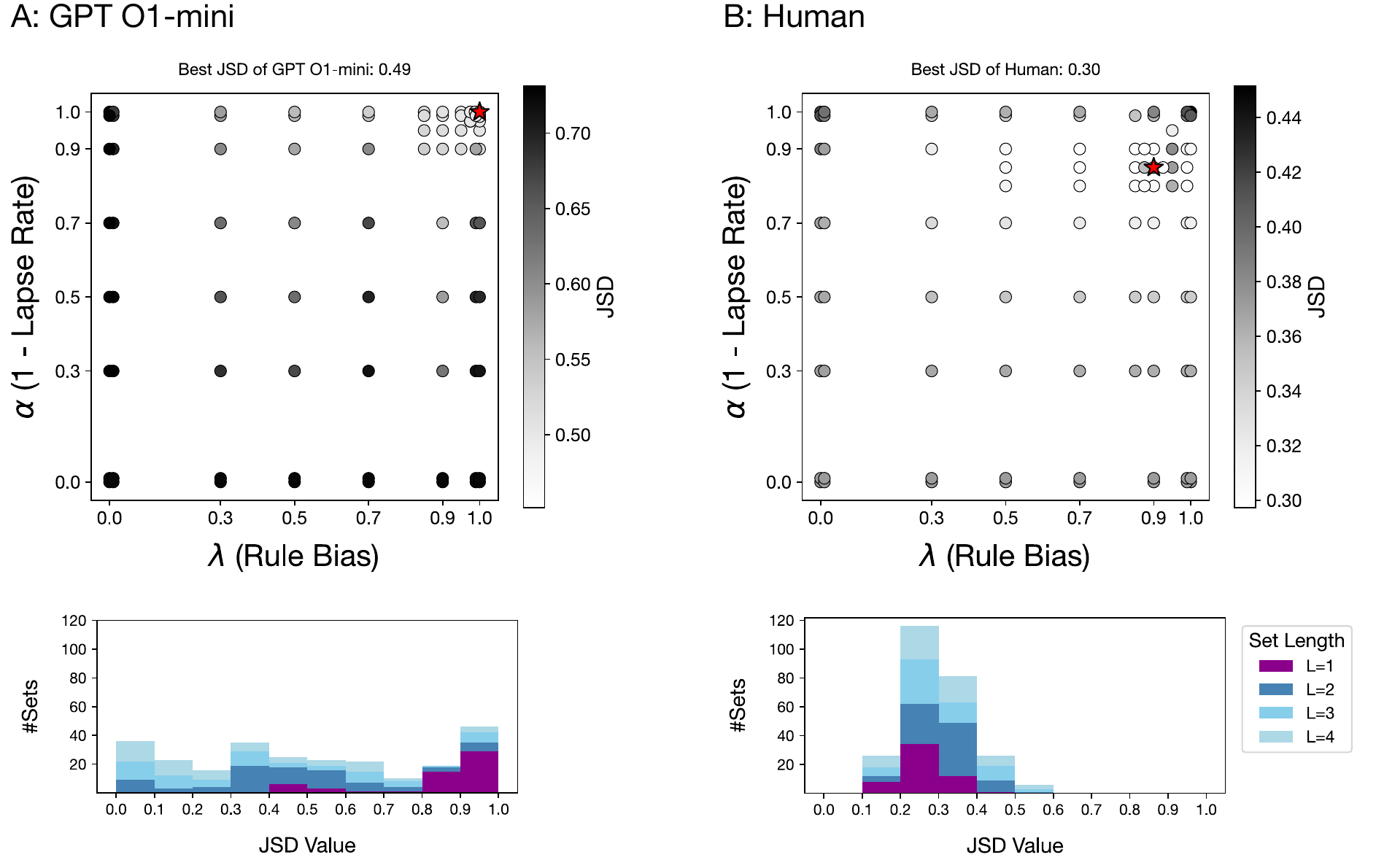}
                \caption{Grid search optimization for $\alpha$ and $\lambda$ parameters. The red star marks the best parameter configuration. Below each panel, histograms show the distribution of per-set JSD values for the best fit.}
                \label{fig:optimization}
            \end{figure}
            \FloatBarrier

\newpage
\subsection{Best model variants} 
\setcounter{figure}{0}
\setcounter{table}{0}
\begin{figure}[htbp]
        \centering
        \includegraphics[width=0.9\linewidth]{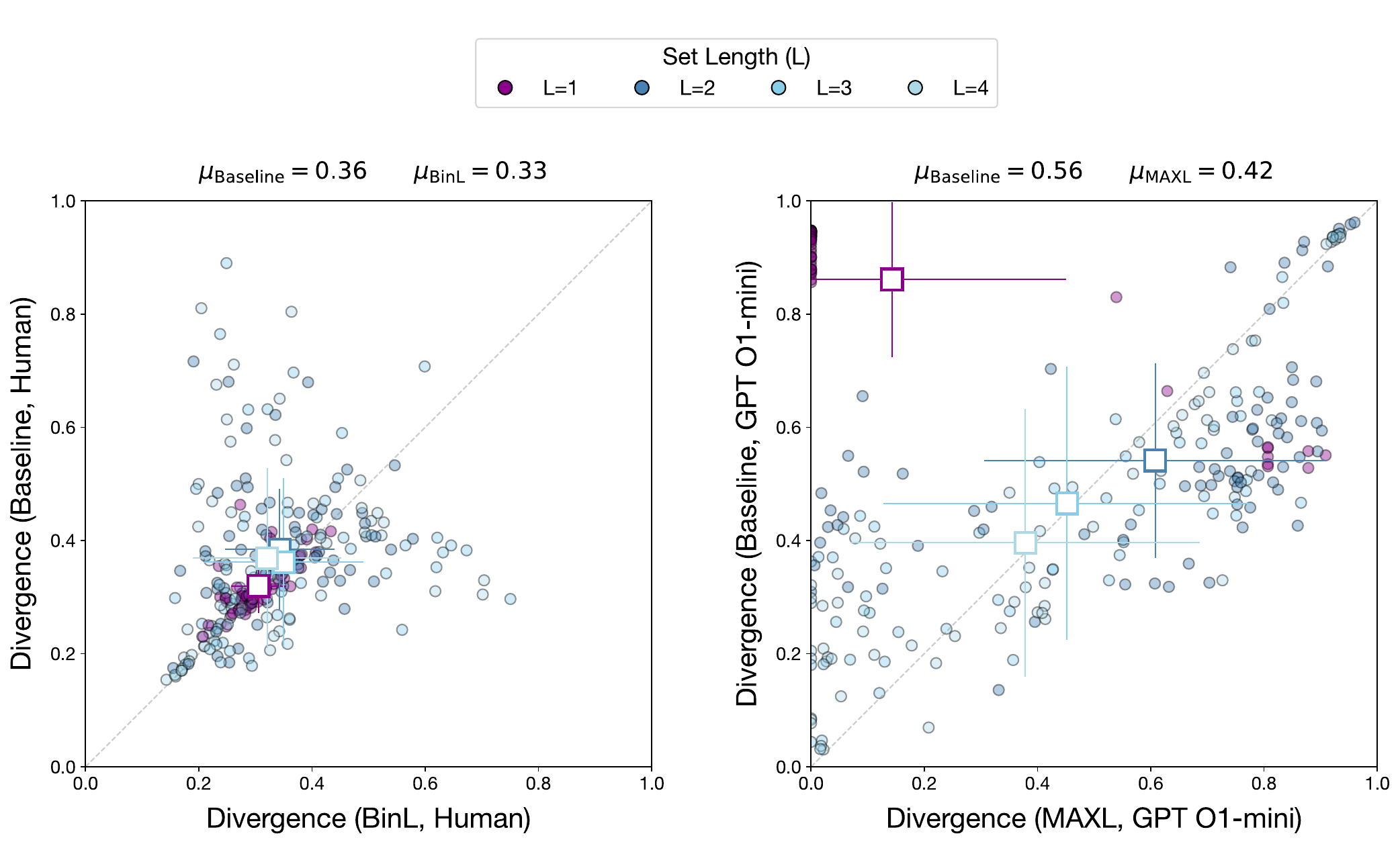}
        \caption{
        Per-set JSD comparisons between each variant (x-axis) and the baseline model (y-axis). Error bars show the standard deviation, clipped at the JSD bounds.
        }
        \label{fig:category_merged}
    \end{figure}
    \FloatBarrier           
\newpage
\subsection{Other LLMs in the number game}
\setcounter{figure}{0}
\setcounter{table}{0}
\begin{figure}[htbp]
            \centering
            \includegraphics[width=0.9\linewidth]{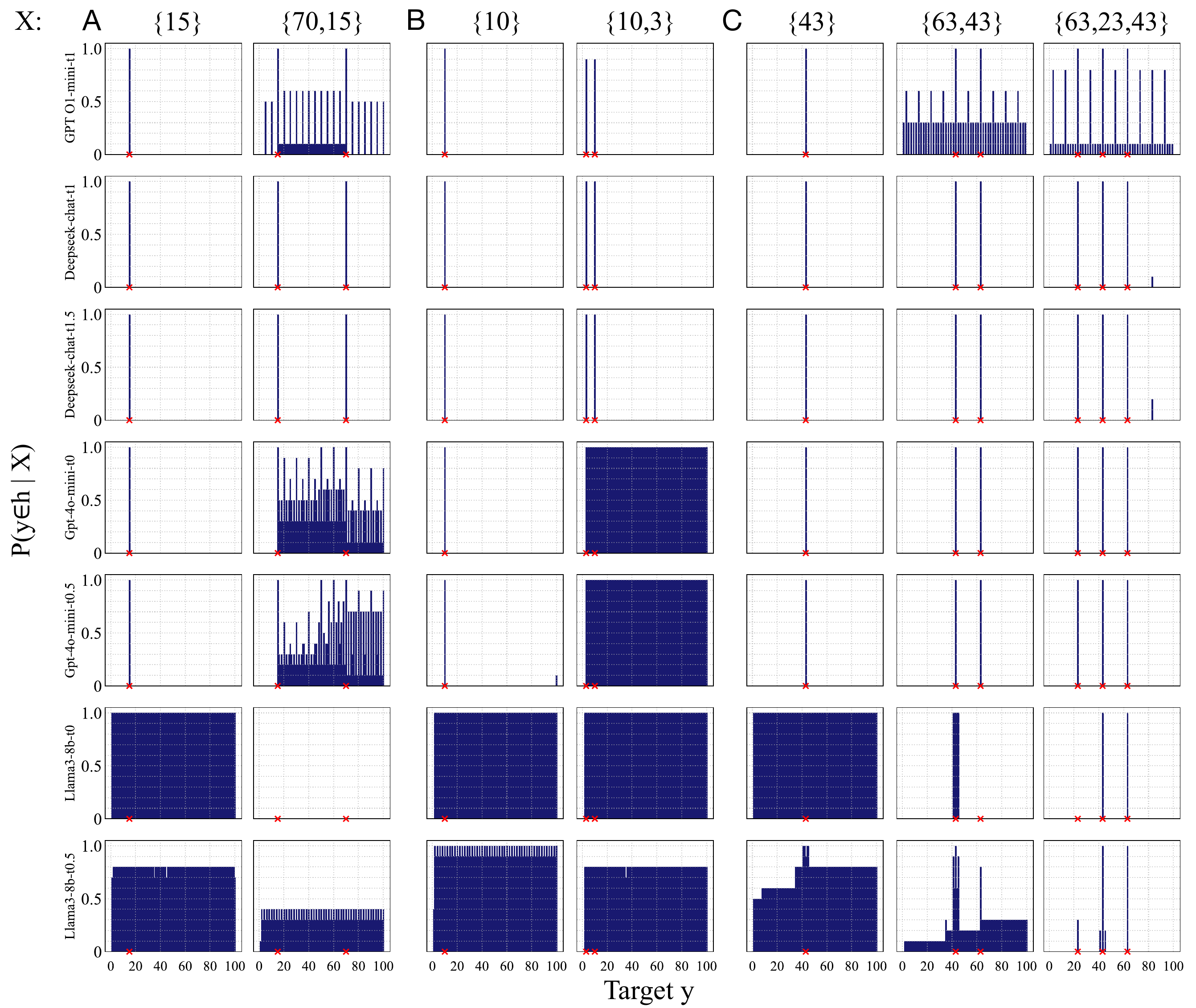}
            \caption{Number-game behavior of other tested LLMs for selected example sets (top titles), showing the probability that a target number $y$ (1-100) belongs to the same concept as the given example set $X$. $t$ indicates the temperature value used in each model.}
            \label{fig:example-llm}
    \end{figure}
    \FloatBarrier

In addition to GPT O1-mini, described in the main text as showing a mixture of rule-based and interval-based generalization, the other models displayed distinct patterns. DeepSeek-V3 largely failed to generalize beyond the given examples. GPT-4o-mini showed some generalization with multiple concepts, with slight variation across temperatures. LLaMA-3-8B often produced saturated responses at temperature 0, rating nearly all numbers very high or failing to select any targets in some cases, but its saturation decreased as the temperature increased.

\end{document}